\documentclass[10pt,twocolumn,letterpaper]{article}

\usepackage{cvpr}              

\usepackage{graphicx}
\usepackage{amsmath}
\usepackage{amssymb}
\usepackage{booktabs}
\usepackage{float}

\usepackage[pagebackref,breaklinks,colorlinks]{hyperref}
\usepackage[accsupp]{axessibility}  

\usepackage[capitalize]{cleveref}
\crefname{section}{Sec.}{Secs.}
\Crefname{section}{Section}{Sections}
\Crefname{table}{Table}{Tables}
\crefname{table}{Tab.}{Tabs.}

\def\cvprPaperID{16} 
\def\confName{CVPR}
\def\confYear{2022}

\begin{document}

\title{VISTA: Vision Transformer enhanced by U-Net and Image Colorfulness Frame Filtration for Automatic Retail Checkout}

\author{Md. Istiak Hossain Shihab\dag\\
Shahjalal University of Science \\ and Technology\\
Sylhet, Bangladesh.\\
{\tt\small istiak@proton.me}
\and
Nazia Tasnim\dag\\
Giga Tech Ltd.\\
Dhaka, Bangladesh.\\
{\tt\small nimzia.blu@gmail.com}
\and
Hasib Zunair\dag\\
Concordia University\\
Montreal, QC, Canada\\
{\tt\small hasibzunair@gmail.com}
\and
Labiba Kanij Rupty\\
Giga Tech Ltd. \\
Dhaka, Bangladesh.\\
{\tt\small labibakanij@gmail.com}
\and
Nabeel Mohammed\thanks{\dag denotes equal contribution.}\\
North South University\\
Dhaka, Bangladesh\\
{\tt\small nabeel.mohammed@northsouth.edu}
}
\maketitle

\begin{abstract}
   Multi-class product counting and recognition identifies product items from images or videos for automated retail checkout. The task is challenging due to the real-world scenario of occlusions where product items overlap, fast movement in conveyor belt, large similarity in overall appearance of the items being scanned, novel products, the negative impact of misidentifying items. Further there is a domain bias between training and test sets, specifically the provided training dataset consists of synthetic images and the test set videos consist of foreign objects such as hands and tray. To address these aforementioned issues, we propose to segment and classify individual frames from a video sequence. The segmentation method consists of a unified single product item- and hand-segmentation followed by entropy masking to address the domain bias problem. The multi-class classification method is based on Vision Transformers (ViT). To identify the frames with target objects, we utilize several image processing methods and propose a custom metric to discard frames not having any product items. Combining all these mechanisms, our best system achieves 3rd place in 
   the AI City Challenge 2022 Track 4 with F1 score of 0.4545. Code will be available at \url{https://github.com/istiakshihab/automated-retail-checkout-aicity22}.
\end{abstract}

\section{Introduction}
\label{sec:intro}
Multi-class product counting and recognition (MPCR) is the task of identifying products from images or videos. Applications of MPCR involve automatic check-out in a store, having huge commercial value in the retail industry. MPCR is challenging due to the real-world scenario of occlusion, movement, similarity in items being scanned, novel products that are created seasonally, and the cost of misdetection and misclassification. Therefore, this naturally becomes an interesting research problem.

\begin{figure}
  \begin{center}
  \includegraphics[width=\linewidth]{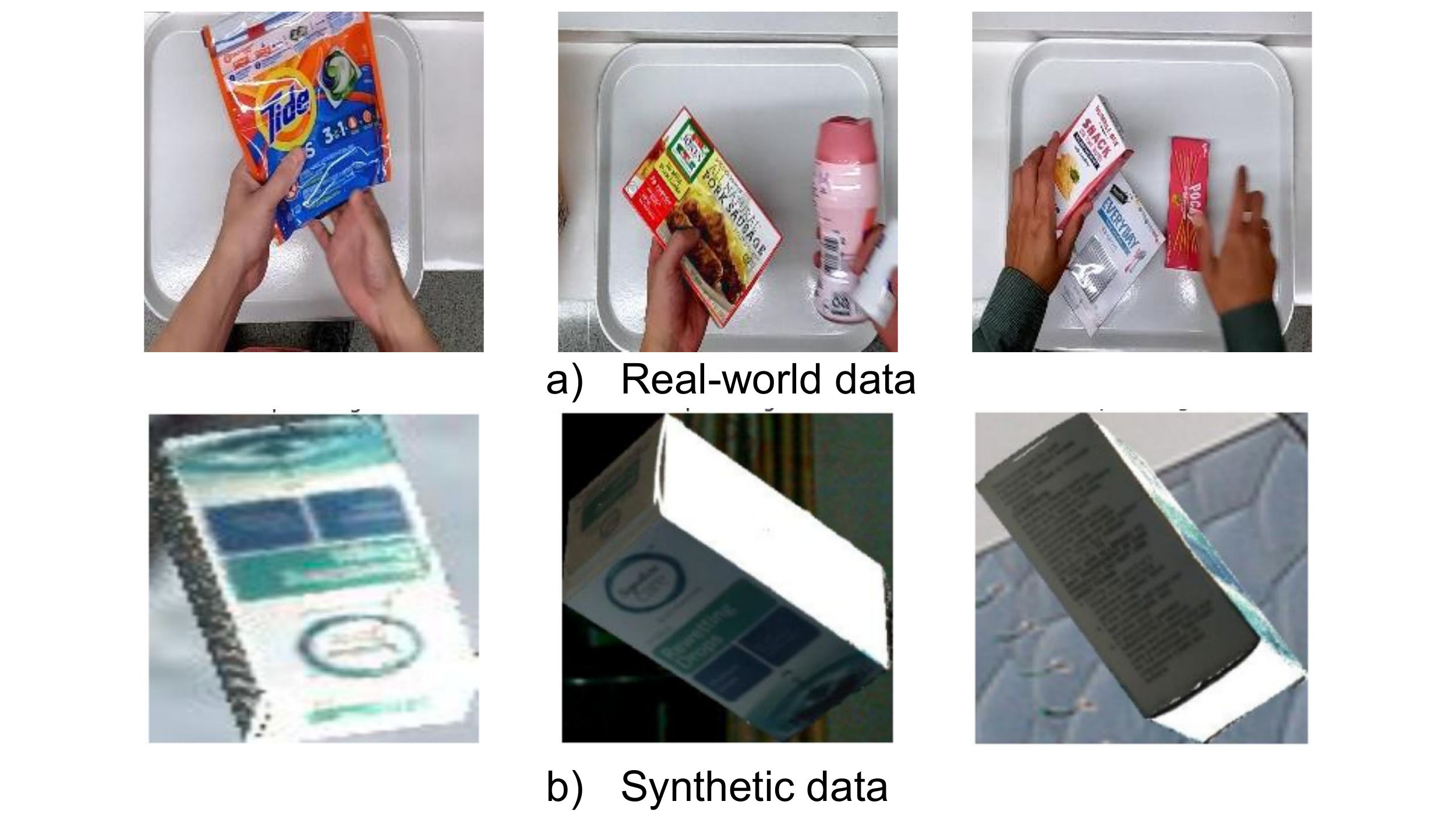}
  \end{center}
      \caption{Example images of the real-world test data and the synthetic data provided for training by AICITY22 Track 4. Notice that in synthetic data there are no foreign objects such as hands and trays.\label{syn}}
\end{figure}

Motivated by the growing applications of machine learning and computer vision, the AI City Challenge 2022 (AICITY22)~\cite{naphade20226th} introduces Track 4: Multi-Class Product Counting \& Recognition for Automated Retail Checkout. As the first version of this new track, the goal is to identify products given a scenario where a customer is hand holding items in front of the checkout counter where the products are moving along a retail checkout tray. Further, the products may be occluded or be very similar to each other. Given a conveyor belt snapshot or video, the goal is to count and identify all products. The task become even more challenging as the provided training set comprises of synthetic images. Performance is tested on a set of objects not included in training for both closed and open-world scenarios. 

To address the MPCR task of AICITY22 Track 4, we, \texttt{The Nabeelians} team, divide it into two sub tasks, segmentation and multi-class classification. Our segmentation stage consists of a single product-, hand- and entropy-segmentation to accurately segment multiple products in an image/video frame. The motivation for hand segmentation lies in the observation that there is a domain bias between training and test data (i.e the training data consists of images with no hands holding the product) as shown in Figure~\ref{syn}. For the classification stage, we propose to use Vision Transformer (ViT)~\cite{dosovitskiy2020vit} as the primary backbone for feature extraction and multi-class image classification. Finally we propose a test frame preprocessing stage that is a combination of different filtration mechanisms and develop a Colorfulness-Binarization-Threshold (CBT) metric that we use to filter out images or frames to segment and then classify.
The main contributions of this paper can be summarized as follows:
\begin{itemize}
\item We propose a segmentation stage consisting of unified single product-, hand- and entropy-segmentation to address the domain bias between training and test data.

\item We propose a classification stage based on Vision Transformers (ViT) for multi-class image classification.

\item We propose a preprocessing stage leveraging custom metric called Colorfulness-Binarization-Threshold (CBT) metric to compute a value to discard images/frames not having any product items.

\item We show that our method achieves third place in the AICITY22 Track 4 final leaderboard results.
\end{itemize}

\section{Related Work}
\label{sec:related}
The adoption of computer vision and machine learning for automating product identification has a great potential in economical and social benefits, mostly because of reliability and saving time. Prior works leverage deep learning for retail product recognition. Specifically, the methods focus on generating data, cross-domain recognition via transfer learning, joint feature learning, incremental learning and regression-based object detection~\cite{wei2020deep}. There are also methods which combine object detection and image retrieval into a single framework for product recognition, given a query and matching it to a reference image of a product item~\cite{tonioni2018deep}. Based on the interest in retail product recognition and scarcity of large-scale datasets for building retailing computer vision systems, Products-10K~\cite{bai2020products} is proposed which consists of 10,000 different product items for fine-grained product recognition. \cite{sinha2022improved} also propose a lightweight approach for easy deployment for automatically identifying each product item. Their approach consists of a Faster-RCNN-based object localizer and a ResNet-18-based image encoder that classifies detected regions into the correct class. Our work is different that prior work in the sense that we are interested in identifying product items (i.e. assign class label) as they pass through a conveyor belt. Note that~\cite{tonioni2018deep,sinha2022improved} works by matching query images with reference images. Also the scenario of~\cite{sinha2022improved} is very different in the sense that they are interested in identifying product items from a given rack image as input by matching it to the query item(s).

\section{Methodology}
\label{sec:method}

\subsection{Problem Formulation}
Given an image frame or video, multi-task product counting and recognition (MPCR) aims to identify product items (i.e assign a class label) from a collection of 116 classes. To solve this task, we are provided with a dataset of \textit{low-}resolution synthetic images of various product item in an image along with a binary segmentation mask indicating the region of interest (ROI) of the product item and the name of the product item. It is important to note that the training set firstly consists of synthetic images. Second, there is only one product item in that synthetic image, while the test set is a video sequence of \textit{high-}resolution where a person is moving the product items and multiple product items appear in the video sequence. Further, the test set videos sequences consists of occlusions, fast movement, large similarity, foreign objects such as hands and tray which make the task very challenging. Finally, we are told to solve an identification and counting problem from a classification viewpoint perspective (i.e. we are given images with the class labels and segmentation masks) which makes it even more difficult as we cannot leverage temporal information of video sequences (i.e. cannot use test data for training for self/semi-supervised learning). For more details, we refer reader to URL.\footnote{https://www.aicitychallenge.org/2022-data-and-evaluation/}

\subsection{Method}

\begin{figure*}
  \begin{center}
  \includegraphics[width=\linewidth]{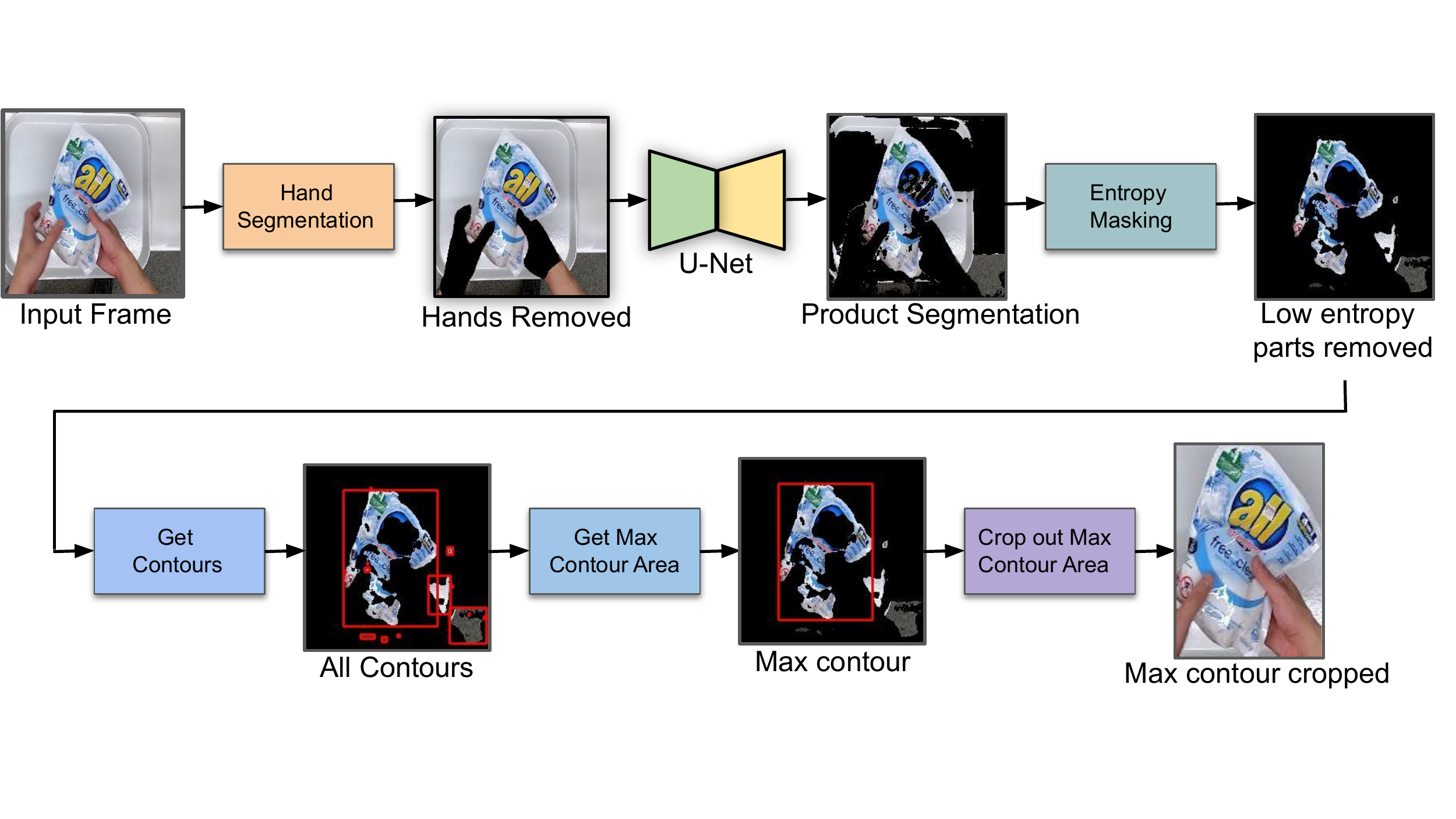}
  \end{center}
      \caption{Schematic layout of the U-Net based segmentation and contour selection stage. Given an optionally filtered input image, we first segment the hand to address the domain bias problem. Then product segmentation is done followed by entropy masking to get the final ROI. After this, contours are located and the maximum contour is kept to reduce noise, which is then cropped from the final image. Finally, the cropped image is fed to ViT, for final classification of the product item.\label{architecture}}
\end{figure*}

\subsubsection{Multi-Product Segmentation}
\medskip\noindent\textbf{Single product segmentation.}\quad Our proposed method is based on the U-Net architecture \cite{ronneberger2015u}, which consists of a contracting path that captures context and a symmetrically expanding path that enables precise localization. In order to localize the upsampled features, the expanding path combines them with high-resolution features from the contracting path via skip-connections \cite{ronneberger2015u}. The output of the model is a pixel-by-pixel binary mask that shows the class of each pixel (i.e. 0 or 1 in our case). 

\medskip\noindent\textbf{Hand segmentation.}\quad To address the domain bias problem, we use out-of-the-box pretrained hand segmentation model~\cite{GuglielmoCamporese} which is a DeepLabV3~\cite{chen2017rethinking} with ResNet-50~\cite{he2016deep} backbone trained on the \texttt{COCO train2017} dataset.

\medskip\noindent\textbf{Entropy masking.}\quad We observed that after removing the hands, there are also other foreign objects present in the frames such as tray and bags on the side etc. We hypothesized that the performance could be further improved if the foreign objects could be removed thus closing the gap between training and test distribution. This motivated us to use entropy masking~\cite{barbieri2011entropy}. The goal is to segregate objects by using the textural cues in the image. After computing the entropy, we binarize the entropy image resulting in a mask. 

\begin{figure}
  \begin{center}
  \includegraphics[width=\linewidth]{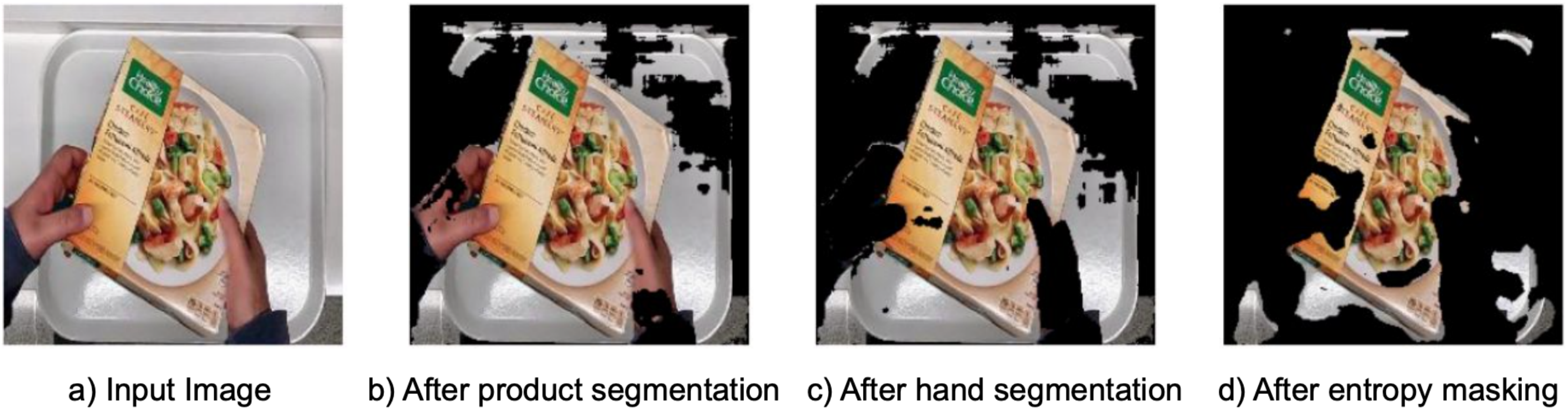}
  \end{center}
      \caption{Example output of product and hand segmentation, followed by entropy masking to get the final ROI, given an image.\label{seg}}
\end{figure}

\subsubsection{Multi-Class Classification}
We have experimented with several architectures in our Multi-Class Classification stage, starting from simple ConvNets~\cite{Bengio95convolutionalnetworks} to gradually more complex and heavy backbones like the Vision Transformers (ViT)~\cite{vit}. However, performance-wise none could match ViT that consistently performed well throughout hyper-parameter tuning process. A summary of the different architectures that we have used including the specific processing steps used before and after the training are summarized in the Table~\ref{tab:performance}.

Below we elaborate on the specifications of our best model configuration and other steps involved.

\medskip\noindent\textbf{Data Preparation.}\quad 
Our training data instances have random background from the \texttt{MS COCO Dataset}~\cite{lin2014microsoft} to diversify the images. However, this is not suitable for our test setting where objects are placed on a rounded-rectangle tray having the color range around white. We preprocess our training dataset to imitate the test instances. The method is elaborated in Section~\ref{sec:expsetup}.

\begin{figure*}[t]
    \centering
    \includegraphics[width=0.9\linewidth]{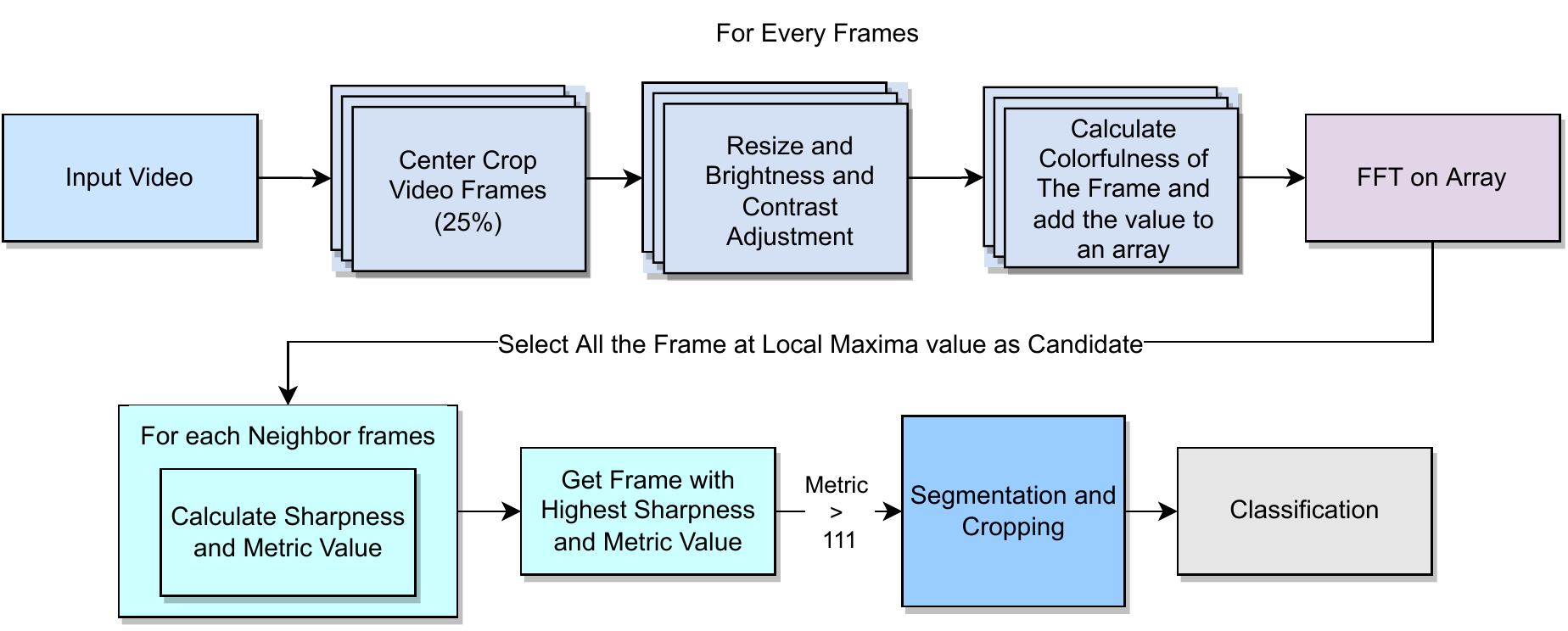}
    \caption{Overview of Test Video Preprocessing and Frame selection stages}
    \label{fig:workflow_diag}
\end{figure*}

\medskip\noindent\textbf{Data Augmentation.}\quad 
Augmentation is a popular data-depended process applied to improve the robustness of models by reducing overfitting. This is particularly suitable when training data is not perfectly suitable for the test setting i.e: lacks diversity. In our case, the training data was a collection of synthetic retail products generated through 3D scanning and the pipeline demonstrated by~\cite{synth}. The objects have random attributes like lighting conditions, background and orientation. Despite its large quantity, the training set has very little similarities with test data, where objects are placed at approximately the center of the frame on a tray by a customer. In contrast to the ideal training instances, there may be multiple objects at once on the tray. The frames are blurry due to fast movement, there's occlusion posed by the presence of customer hands. Random objects may be present nearby the region of interest (ROI) and lighting condition may change instantly. Thus to mitigate these issues we utilize data augmentation in our training pipeline. 
\begin{itemize}
    \item We begin with RandAugment ~\cite{randaug}, an automated data augmentation policy that uniformly samples operations from a set of augmentations - such as equalisation, rotation, solarization, colour jittering, posterizing, changing contrast, changing brightness, changing sharpness, shearing, and translations - and sequentially applies a number of these.  For our training, we use it with a magnitude of 7 and 0.5 standard deviation of noise. 
    \item CutMix~\cite{cutmix} improves a model's localisation capability by forcing it to recognise an item from a partial view. This is particularly suitable for our task, because test data may appear in variety of orientations and have portions of it occluded by hands or other items.
    \item MixUp ~\cite{mixup} augmentation improves robustness and it's extremely good at regularization of DNN models. For training we used a mixup alpha of 0.3. 
\end{itemize}

\medskip\noindent\textbf{Vision Transformer.}\quad 
The backbone of our final classification model is (ViT)~\cite{vit}. ViT divides an image into fixed-size patches, linearly embeds each of them, adds positional embeddings, and feeds the resulting vector sequence to a standard Transformer encoder~\cite{encoder}. A linear layer or MLP head is used for classification. We have used the ViT Base pretrained on the
\texttt{ImageNet-21k}~\cite{imnet} dataset at $224\times224$ resolution and patch dimension $32\times32$ pixels. 

\subsubsection{Test-set Pre-Processing and Filtration}
\label{preproc}

As described in Section~\ref{results}, our test set consists of $5$ videos of on average $30$ seconds duration with 60 frames per second. We separate these videos into frames using OpenCV~\cite{opencv_library} and apply several processing functions on them to pick the most interesting frames. In this section, we discuss the different process the test frames go through in order to get picked for classification. A high-level overview of these filtration stages are shown in Figure~\ref{fig:workflow_diag}.

\medskip\noindent\textbf{Frames to ROI.}\quad As discussed earlier, all the frames from a test video is first extracted. This is followed by a 25\% crop of these frames. The cropping percentage is obtained through manual experimentation. After this, we get an approximate estimation of the ROI. We intentionally crop parts outside of tray to have a safe estimation in case the product er placed on side of the tray. For each frame, the brightness and contrast is adjusted and the image resized to $224\times224$. 

\begin{figure}[t!]
    \centering
    \includegraphics[width=0.9\columnwidth]{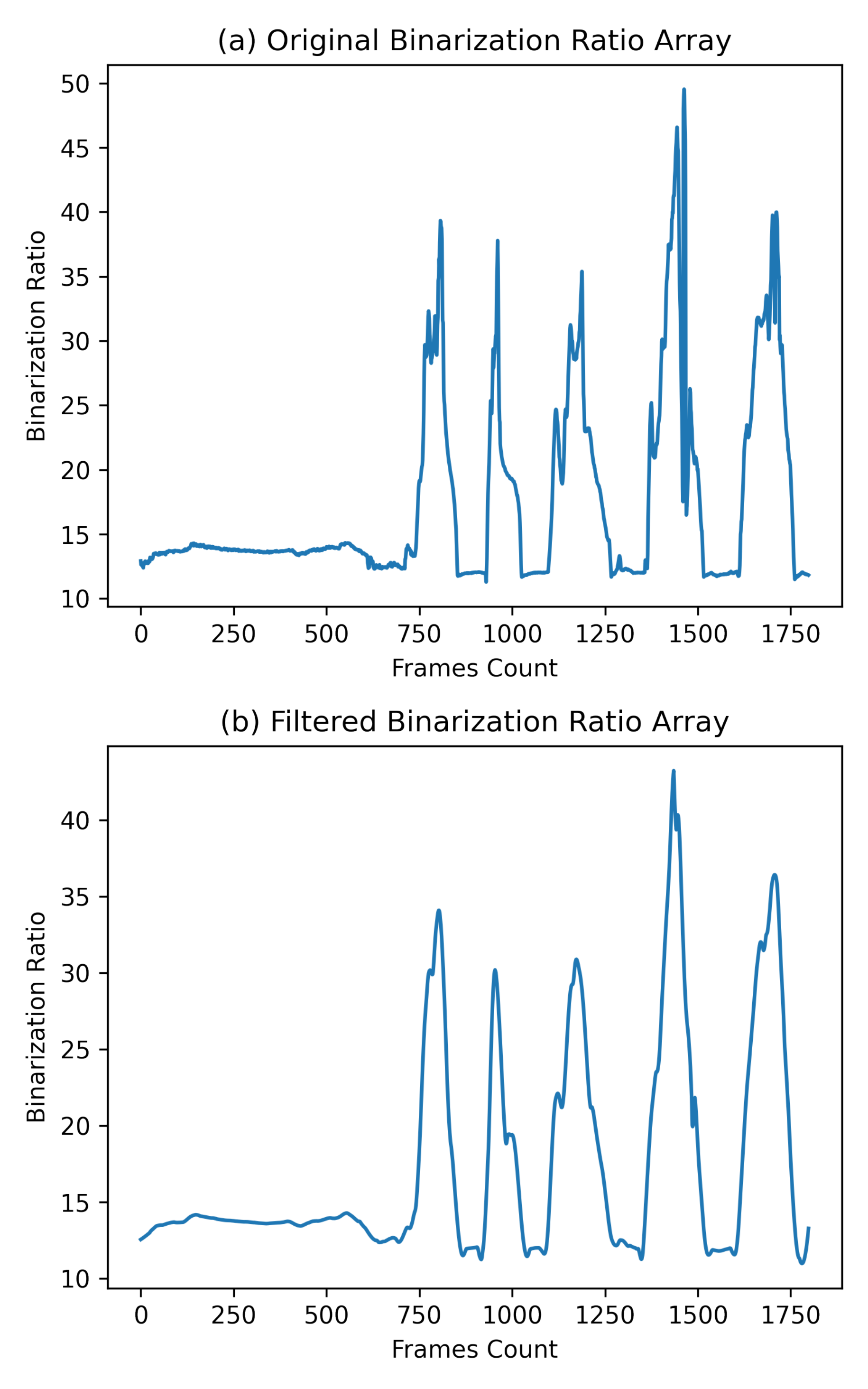}
    \caption{Local maxima determination after smoothing binarization ratio array}
    \label{fig:savgol-filter}
\end{figure}

\begin{figure*}[ht!]
    \centering
    \includegraphics[width=\textwidth]{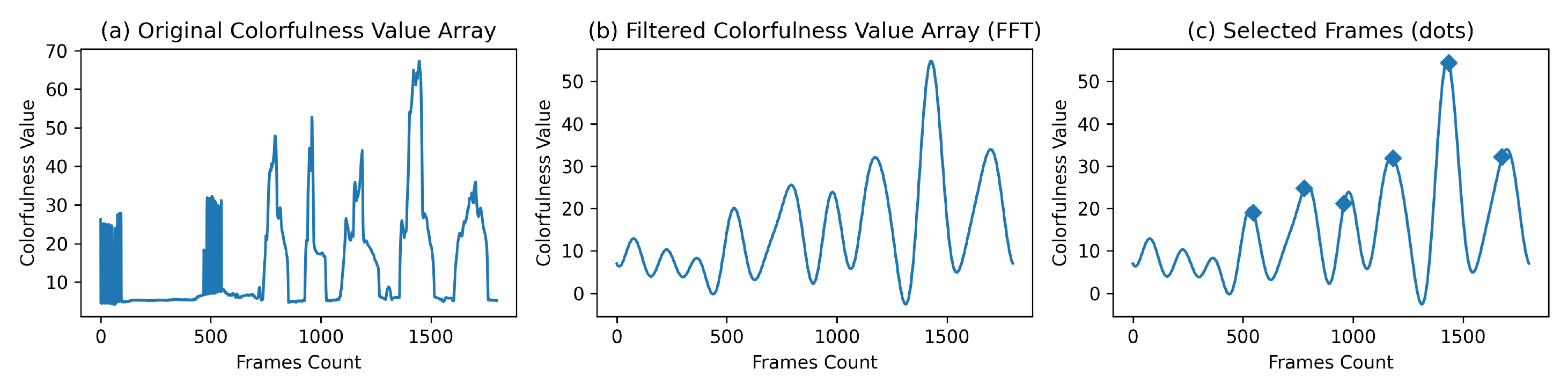}
    \caption{Local maxima determination after smoothing colorfulness ratio array}
    \label{fig:colorfulnessfft}
\end{figure*}

\medskip\noindent\textbf{Frame Selection through Binarization.}\quad
For each frame of our test video, we first convert them to grayscale and apply a binary inverse thresholding. The same threshold value is used for each pixel. As described by  equation \ref{bin_inv}, if the pixel value is less than the threshold, it is set to 0, otherwise a maximum value is used. In the equation, $src$ and $dst$ refer to initial and final value for point $(x,y)$ respectively.

\begin{equation}
\label{bin_inv}
    dst(x,y) =  \left\{\begin{matrix}
0 & if, src(x,y) > thresh\\ 
maxval & otherwise
\end{matrix}\right.
\end{equation}

We use OTSU Binarization~\cite{otsu_binarization} to determine the optimal global threshold value from the image histogram and it is denoted by $thresh$ in equation~\ref{bin_inv}. For the binarized image we count the number of non-zero pixels, and calculate its ratio percentage against the whole image. The calculated ratio for each frame is stored inside an array, which is input to the Savitzky-Golay filter\cite{savitzky1964smoothing} for smoothing. We pick the local maxima from this smoothed array as our candidate frames with objects. Figure ~\ref{fig:savgol-filter} shows the transformation of raw array to filtered array with cleaner maxima.

\medskip\noindent\textbf{Sharpness Calculation of cadidate frames.}\quad
After frame selection step (colorfulness-based), we can refine the selections based on their sharpness. The objective is to choose the sharpest image among the almost similar frames for the same object, to make the classification task more efficient. To do this, we pick 7 frames around the local maxima, each at 7 timesteps away from the other. For each of these 7 frames, we calculate the sharpness by estimating the average gradient magnitude ~\cite{birdal_2022}. 

\medskip\noindent\textbf{Frame Selection based on Image Colorfulness}\quad 
Colorfulness of image is calculated based on a metric developed by Hasler and Suestrunk~\cite{colorfulness}, where the authors use the idea of opponent color space representations along with the mean and standard deviations of these representation values. The metric is defined in Figure \ref{eq:colorfulness}, where $R$, $G$ and $B$ denotes each color space. $rg$ and $yb$ denotes  opponent color spaces. $\sigma$ denotes Standard Deviation and $\mu$ denotes Mean value. 

\begin{eqnarray}
\label{eq:colorfulness}
  \left\{
  \begin{aligned}
    &rg = R - G \\
    &yb = \frac{1}{2}\left ( R +G  \right ) - B \\
    &\sigma _{rgyb} = \sqrt{\sigma ^2 _{rg} + \sigma ^2 _{yb} } \\
    &\mu _{rgyb} = \sqrt{\mu ^2 _{rg} + \mu ^2 _{yb}}\\
    &colorfulness = \sigma _{rgyb} +  0.3*\mu _{rgyb}
  \end{aligned}
  \right.
\end{eqnarray}

We use it to calculate the colorfulness of each test frames to a list. FFT is applied on this list to smoothen it, and similar to the \textit{Frame Selection through Binarization} process, we identify the local maxima as our candidate frames which we assume to have objects in them.  The transformation of raw array to filtered array with cleaner maxima is presented in Figure~\ref{fig:colorfulnessfft}. We indicate our candidate frames using dots(.) on the plot. 

\medskip\noindent\textbf{CBT Metric to Identify Final Frames.}\quad
For the colorfulness-based frames that were further filtered through the steps mentioned earlier, we additionally calculate their binarization ratio as discussed in $Frame Selection through Binarization$ subsection. We use the amount of colorfulness with this binarization ratio to compute a custom metric value. We name this metric as Colorfulness-Binarization-Threshold (CBT) metric and define it in equation~\ref{eq:IC}. : 

\begin{equation}
  CBT\_metric\_value = \sqrt{c\_ratio^2 \times b\_ratio}
  \label{eq:IC}
\end{equation}
 Here, $c\_ratio$ is the colorfulness ratio and $b\_ratio$ is the binarization ratio. We set the threshold for the sharpness metric value to 111 and only keep a frame if it has has $sharpness\_value > 111$. This threshold value was picked through manual experimentation. This metric has been derived based on the observation that, colorfulness ratio has significant contribution in selecting the better frames. While, binarization ratio does not have the same effect, it still helps in the detection. 

\medskip\noindent\textbf{Contour Selection and Duplicate Frames Removal.}\quad After obtaining the selected frames from the earlier steps,  we run product and hand segmentation followed by entropy masking to get the masked outputs and find the contours from it. The contour with the maximum area is selected and cropped by a bounding box position from the original frame. We also experiment in this stage, by first calculating Root Mean Square (RMS) of the contour areas. This experiment is motivated by trying to capture multiple objects having higher contour areas than the RMS value. The contour selection step is followed by multi-class classification which assigns a class label to the frame from among the 116 different classes.

Lastly, we check if consecutive frames are part of the same video and the detected objects have appeared before within a close time interval. If that is the case, we discard the latter frame with prediction, otherwise they are kept as distinct detection.





\subsubsection{Training}
\medskip\noindent\textbf{Segmentation.}\quad Given the synthetic images and their segmentation labels, we train a U-Net architecture \cite{ronneberger2015u} using Adam solver \cite{kingma2014adam} to minimize the pixel-wise binary cross-entropy loss function with a learning rate 0.0001 and a batch size of 16. Training is done for 10 epochs or until the validation loss stagnates using an early stopping
mechanism, and then the best weights are retained.

\medskip\noindent\textbf{Classification.}\quad 
The input to our classification stage is the background removed and augmented object dataset and their associated labels. We train our model by optimizing the Categorical Cross-entropy loss through the \texttt{lookahead AdamW}~\cite{adamw} function. For all our trained models, we have used dropout rate $0.05$, learning rate $0.005$, weight decay rate $0.01$ and batch size of $32$ . We train our models for $50$ epochs with an early stopping patience of $5$, that only saves the best model.
\subsubsection{Inference}
\medskip\noindent\textbf{Segmentation.}\quad Given an image, we first segment the product item using the trained U-Net~\cite{ronneberger2015u} which is followed by hand segmentation~\cite{GuglielmoCamporese}. Then, we perform entropy masking to get the final ROI (i.e. product item). We show a depiction of the process in Figure~\ref{seg}.

\medskip\noindent\textbf{Classification.}\quad
For our earlier models, we directly use the default dataset for training and perform a very simple black and white or binarized color ratio-based pre-inference processing elaborated in Section~\ref{preproc}, on the test video frames. These models couldn't score over $15\%$ on the leaderboard. 

We then move to using the background-replaced dataset (BRD) and focus on improving the frame selection process. We shift from using binarized color ratio to quantifying colorfulness in image, as frame selection criteria. Per 6 consecutive frames, we calculate the amount of blurriness and pick the sharpest among them. This whole mechanism is also extensively discussed in Section~\ref{preproc} . For the sake of convenience, we will denote this mechanism as $Frame\_selection\_color$ and its predecessor as $Frame\_selection\_binarized$. With the $Frame\_selection\_color$ process we progressively attach our segmentation pipeline, a contour selection step and a duplicate removal step.

\begin{table*}[ht]
   \begin{center}
   \begin{tabular}{|c|p{10cm}|c|}
   \hline
   \textbf{Model} & \textbf{Pre-inference Processing} & \textbf{F1 Score} \\
   \hline\hline
   Baseline EffNet-B0 & Default Dataset + Frame\_selection\_binarized & 0.14 \\
   MobileNet-V3 & BRD + Frame\_selection\_binarized  & 0.037 \\
   ViT\_base & BRD + Frame selection binarized   & 0.14 \\
   ViT\_base & BRD + Frame\_selection\_color + CBT Metric & 0.29 \\
   ViT\_base & BRD + Frame\_selection\_color + Segmentation and Contour Selection (Max) + CBT Metric & 0.4255 \\
   \textbf{ViT\_base} & \textbf{BRD + Frame\_selection\_color  +  Segmentation and Contour Selection (Max) + Duplicate Frame Removal + CBT Metric} & \textbf{0.4545} \\
   ViT\_base & BRD + Frame\_selection\_color  +  Segmentation and Contour Selection (RMS) + Duplicate Frame Removal + CBT Metric & 0.42 \\
   ViT\_base & BRD + Frame\_selection\_color +  Segmentation and Contour Selection (Max) + Duplicate Frame Removal + Crop Re-segmentation + Contour Selection (Max) + CBT Metric  & 0.4444 \\
   \hline
   \end{tabular}%
   \end{center}
   \caption{Ablation study results on the test set of AICITY22 Track 4. Boldface indicates our best model. 
   \label{tab:performance}}
\end{table*}

\section{Experiments}
\label{sec:results}

\subsection{Experimental setup}
\label{sec:expsetup}
\medskip\noindent\textbf{Dataset details.}\quad The AICITY22 Track 4 training dataset contains a total of $116,500$ synthetic images and  their corresponding segmentation labels of $116$ different retail merchandises. After scanning an item, 3D models are created to generate synthetic pictures.  Synthetic data is used in this case because it can be used to create large-scale training sets in a variety of settings, including different lighting conditions, orientations, noisy surroundings, and random movements. It is mention worthy that the dataset is balanced across all 116 different classes with on average 900 images per label.

The camera is set above the checkout counter and staring straight down in the test scenario, while a customer performs a checkout by "scanning" things in front of the counter in a realistic manner. To add to the intricacy, several distinct clients participated in the checkout process, each of whom scanned somewhat differently. A shopping tray is placed beneath the camera to highlight the model's attention point. Customers who participate may or may not place items on the tray. Several full scanning activities involving one or more objects are contained in a single video clip. The test video lengths are on average 30 seconds, having a frame rate of 60 and resolution $1920\times1080$. 


\medskip\noindent\textbf{Dataset challenges.}\quad
The provided training and test dataset were intrinsically challenging for both training and testing sets. In reality, the dataset is better suited for classification tasks, but for this track we had to perform both object localization and counting with classification in the test phase. While solving this task, we were careful about the various obstacles and took actions, accordingly. 
We enumerate the data challenges as follows: 
\begin{itemize}
    \item The training data had no real resemblance with the testing data as they were 3D simulated, for the sake of diversity. The test data is video sequences of real-time retail object checkout performed by several different customers.
    \item Training data were very jittery and blurry, due to the scanning. No title or text could be deciphered from them. However, despite having a clear appearance the test data consisted of objects occluded by foreign objects such as hands and tray. The test objects were also constantly in motion.
    \item Training data were zoomed to exactly crop around the object border and had no noisy information in the frame and were of low-resolution. Our testing video sequences were firstly of high-resolution and comparatively crowded with nearby surroundings that changes often in terms of lighting conditions, camera angles and object orientations. 
    \item Each training sample had only one object, which is contrary to much of our test frames where multiple object of varying size, color and label were present. We saw at most four or five objects in a single frame. 
\end{itemize}

\medskip\noindent\textbf{Data preprocessing.}\quad 
To make the train set similar to the test images, we use the segmentation labels of the train objects and use it to mask out the object's complex background. The background is then randomly replaced with a rectangular or circular gradient scene. 
To generate this gradient background, we first define an inner color and an outer color. Then a distance-based ratio for each $(x,y)$ pixel is calculated. Finally, we enumerate the color value for that point using the following formula :
\begin{equation*}
color_{xy} = (inner\_color * ratio) + (outer\_color * (1 - ratio))
\end{equation*}

\medskip\noindent\textbf{Implementation details.}\quad Experiments are conducted using Python programming language and PyTorch deep learning framework. A full training takes roughly 5-6 hours. Experiments are performed on a workstation with dual NVIDIA RTX 2080Ti GPUs and another workstation with single NVIDIA RTX 3090.

\medskip\noindent\textbf{Evaluation metrics.}\quad 
For this task, the F1-score is the primary metric used to evaluate performance in terms of model identification. Additionally, precision and recall is also calculated. These metrics are calculated based on the following definitions :
\begin{itemize}
    \item \texttt{True Positive} is considered when an object is correctly idetified inside the region of interest and within the appropriate time duration in the video.
    \item \texttt{False positive} is when an object is identified but it is either incorrect or not within the correct time duration. 
    \item \texttt{False Negative} is when an object is not identified, despite being present in a frame. 
\end{itemize}

A high precision tells us that, even if the model can not find all the objects in a frame, but those it can identify are usually correct. Alternatively, a high recall hints that the model can identify most of the objects but it may label some of them wrongly. F1 score is defined as the harmonic mean of precision and recall, and for multi-class and multi-label case, this is the average of the F1 score of each class with weighting. It is also called $macro-F1$ score~\cite{1911.03347} and it is given by equation ~\ref{eq:macro_f1}. Here $pj$ and $rj$ denotes precision and recall for each class, respectively and $Q$ denotes the number of classes.

\begin{equation}
\label{eq:macro_f1}
    Macro F1 = \frac{1}{Q} \sum_{j=1}^{Q}\frac{2*pj*rj}{pj+rj}
\end{equation}

\begin{table}
   \begin{center}
   \begin{tabular}{|l|c|c|c|}
   \hline
   \textbf{Rank} & \textbf{Team ID} & \textbf{Team Name} & \textbf{F1 Score} \\
   \hline\hline
   1 & 16 & BUPT-MCPRL2 & 1.0 \\
   2 & 94 & SKKU Automation Lab & 0.4783 \\
   \textbf{3} & \textbf{104} & \textbf{The Nabeelians} & \textbf{0.4545} \\
   4 & 165 & mizzou & 0.4400 \\
   5 & 66 & RongRongXue & 0.4314 \\
   6 & 76 & Starwar & 0.4231 \\
   7 & 117 & GRAPH\@FIT & 0.4167 \\
   8 & 4 & HCMIU-CVIP & 0.4082 \\
   9 & 9 & CyberCore-Track4 & 0.4000 \\
   10 & 55 & UTE-AI & 0.4000 \\
   11 & 160 & KiteMetric & 0.3929 \\
   12 & 32 & AICLUB\@UIT & 	0.3922 \\
   13 & 163 & Titans-UTE-AI & 0.3774 \\
   14 & 112 & mt\_vacv & 0.3404 \\
   15 & 49 & Sertis & 0.3404 \\
   16 & 170 & PanxUofg & 0.3396 \\
   \hline
   \end{tabular}
   \end{center}
   \caption{Overall F1 score achieved by different methods on the final test set by AICITY22 Track 4. Results taken from final leaderboard. Boldface indicates our approach and ranking.\label{results}}
\end{table}

\subsection{Results}

\medskip\noindent\textbf{Ablation Study.}\quad 
In Table~\ref{tab:performance}, we show results on the test set of AICITY22 Track 4 made during the challenge. We can see that ViT, U-Net and CBT metric based approaches leads to the best results. We can see that the base ViT model has a relative improvement of 107.1\% which is significantly better that the Baseline EfficientNet-B0~\cite{tan2019efficientnet} model. Despite achieving the same F1 score at final test set, while experimenting on the then released test set, ViT scored 10 points higher than Efficientnet-B0 on same configuration. So, we didn't pursue EfficientNet any further on CBT metric. Next, after segmentation and contour selection we see a relative improvement of 44.8\%. Performance improvement is also observed after removing the duplicate frames. Finally, using the CBT metric in combination of other test set preprocessing steps, we get our best model which has a F1 score of $0.4545$.

\medskip\noindent\textbf{Leaderboard results.}\quad Table~\ref{results} shows the final rankings of the different methods using F1 score as the primary measure of evaluation in the AICITY22 Track 4 leaderboard. Our approach achieves 3rd place overall by a good margin.

\section{Conclusion}
\label{sec:conclusion}
We propose a two-stage segmentation and classification framework for identifying product items from video frames for automated retail checkout. Our segmentation system is a unified
single product item and hand segmentation stage followed by entropy masking to address the domain bias problem. After this we use a Vision Transformer (ViT) for classification. In this paper, we also demonstrate the effects of several metrics to identify the best frames from the vastly different test set, to determine the correct region of interest and removing unnecessary noise from the frames. To do this we also create a new metric (CBT metric) that serves our purpose for this dataset. Our best method achieves \textbf{3rd place} in the AI City Challenge 2022 Track 4. For future work, we aim to exploit the temporal information in video frames for better identification of product items. We would also continue our experiments on other architectures and hyperparamters, as well as augmentation techniques. 

\newpage
{\small
\bibliographystyle{ieee_fullname}
\bibliography{references}
}

\end{document}